\documentclass[12pt,a4paper]{article}

\usepackage[T1]{fontenc}
\usepackage[utf8]{inputenc}
\usepackage[protrusion=true,expansion=false]{microtype}

\usepackage[a4paper,margin=2.5cm]{geometry}
\usepackage{setspace}
\usepackage{amsmath,amssymb,amsfonts}

\usepackage{graphicx}
\usepackage{booktabs}
\usepackage{array}
\usepackage{multirow}
\usepackage{xcolor}
\usepackage{caption}
\usepackage{enumitem}

\usepackage[round,authoryear]{natbib}
\setcitestyle{aysep={}}

\usepackage[colorlinks=true,linkcolor=blue!50!black,citecolor=blue!50!black,
            urlcolor=blue!50!black,plainpages=false,hypertexnames=false]{hyperref}
\usepackage{url}
\usepackage{etoolbox}
\apptocmd{\thebibliography}{\raggedright}{}{}
\providecommand{\doi}{}
\renewcommand{\doi}[1]{\href{https://doi.org/#1}{\nolinkurl{doi:#1}}}
\usepackage{titlesec}
\titleformat{\section}{\normalfont\large\bfseries}{\thesection.}{0.6em}{}
\titleformat{\subsection}{\normalfont\normalsize\bfseries}{\thesubsection.}{0.6em}{}
\titlespacing*{\section}{0pt}{1.4\baselineskip}{0.4\baselineskip}

\begin{document}

% =====================================================================
\begin{titlepage}
\begin{center}
\vspace*{0.5cm}
{\LARGE\bfseries The System Hallucination Scale (SHS): A Human-Centered Instrument for Assessing Perceived Reliability of Large Language Models\par}

\vspace{1.4cm}

{\large
Heimo M\"uller\textsuperscript{a,b},\quad
Dominik Steiger\textsuperscript{c},\quad
Markus Plass\textsuperscript{a},\quad
\\Andreas Holzinger\textsuperscript{a,d,*}
\par}

\vspace{1.0cm}

{\small
\textsuperscript{a}\,Machine Learning and Information Science Group, Medical University of Graz, Graz, Austria\\[2pt]
E-mail: heimo.mueller@medunigraz.at\\
E-mail: andreas.holzinger@medunigraz.at\\
E-mail: markus.plass@medunigraz.at\\
\textsuperscript{b}\,Human Machine Mind Cooperation, Graz, Austria\\[2pt]
E-mail: heimo.mueller@hmmc.at\\
\textsuperscript{c}\,MIDATA Cooperative, Zurich, Switzerland\\[2pt]
E-mail: dominik.steiger@midata.coop\\
\textsuperscript{d}\,Human-Centered AI Lab, BOKU University Vienna, Vienna, Austria\\[10pt]
\textsuperscript{*}\,Corresponding author. Andreas Holzinger, Human-Centered AI Lab, FTEC, Department of Ecosystem Management, Climate and Biodiversity, BOKU University, Vienna, Austria.\\
E-mail: andreas.holzinger@boku.ac.at
\par}
\vfill
Note: A preliminary version of this paper is available at the arXiv preprint server:

Müller, H., Steiger, D., Plass, M., \& Holzinger, A. (2026). The System Hallucination Scale (SHS): A Minimal yet Effective Human-Centered Instrument for Evaluating Hallucination-Related Behavior in Large Language Models. arXiv preprint arXiv:2603.09989.
Link: https://arxiv.org/abs/2603.09989

\vfill

\end{center}
\end{titlepage}

% =====================================================================
% TITLE (running) + ABSTRACT
% =====================================================================
\begin{center}
{\large\bfseries The System Hallucination Scale (SHS): A Human-Centered Instrument for Assessing Perceived Reliability of Large Language Models\par}
\end{center}

\vspace{0.5cm}

\noindent\textbf{Abstract}\\[4pt]
Large language models (LLMs) routinely produce fluent text that is factually
wrong, weakly grounded, or misleading---behavior commonly termed
\emph{hallucination}. Evaluation of this behavior is dominated by automated
benchmarks that quantify system-internal correctness but are silent on how
users \emph{perceive} reliability while interacting with a system. We introduce
the \textbf{System Hallucination Scale (SHS)}, a lightweight instrument that
measures \emph{perceived reliability} of LLM output along five interpretable
dimensions: factual accuracy, source reliability, logical coherence,
deceptiveness of presentation, and responsiveness to user guidance. Following
the design tradition of the System Usability Scale and the System Causability
Scale, SHS uses ten paired positive/negative items on a five-point Likert
scale, yielding a directional score and a built-in consistency diagnostic. In a
\emph{supervised validation study}---a protocol-structured,
hallucination-eliciting session with participant-composed prompts, not a
naturalistic field deployment---47 trained experimenters
administered the instrument to 193 participants (386 LLM evaluations, each
followed by a full SHS administration). SHS showed high perceived
clarity (87.2\% of administrators), strong internal consistency (Cronbach's
$\alpha = .87$, 95\% CI $[.84, .90]$), and theoretically coherent inter-dimension correlations
($r = .42$--$.72$, all $p < .001$). The observed dimension profile is itself
diagnostic: under a common two-category prompt protocol, participants rated
source verifiability
($s = 0.18$) and non-deceptive presentation ($s = 0.09$) markedly lower than
factual accuracy ($s = 0.34$), locating the perceived weakness in
\emph{traceability and framing} rather than in raw correctness. The instrument
is fast to complete (mean 4.2 min), requires minimal instruction, and is
released in three languages with open reference code and an interactive
calculator. We are deliberate about scope: this is a proof of concept for the
instrument, establishing that ten SHS items behave as SUS and SCS items do. It
is not a hallucination detector, not an objective hallucination study, not a
model benchmark, and not a naturalistic field deployment; SHS measures
perceived, not ground-truth, reliability, and has not yet been correlated with
any objective hallucination measure. 

\vspace{0.4cm}
\noindent\textbf{Keywords:} human-centered AI; large language models;
hallucination; perceived reliability; psychometric evaluation;
usability instruments

\vspace{0.6cm}

% =====================================================================
\section{Introduction}
% =====================================================================

Interactive computing has entered a phase in which the conversational partner
is itself a statistical model of language. Large language models (LLMs) now
mediate everyday tasks such as search, writing, tutoring, and decision support,
and are increasingly accessed through mobile apps, browser assistants, and
embedded conversational agents \citep{Brown:2020:LLM}. As they move from
laboratory benchmarks into the systems and contexts in which people actually
work, communicate, and seek information, a human-factors problem comes to the
fore: LLMs frequently produce text that is fluent and persuasive yet factually
incorrect, weakly grounded, or fabricated \citep{Zhou:2024:Reliability}. This
phenomenon---\emph{hallucination}---is not merely a technical defect of the
model; it is an \emph{interaction} problem, because its consequences depend on
how a human reads, trusts, and acts upon the output
\citep{Si:2024:Verify,Kim:2024:Uncertainty}.

Despite broad recognition of the problem, evaluation has largely continued a
long-standing tradition in computer science of optimizing quantifiable,
system-level indicators---accuracy, benchmark scores, and efficiency
\citep{Chang:2024:SurveyLLM}. These metrics are indispensable for model
development, but they reduce a rich socio-technical situation to a single
correctness number and say little about the dimensions that matter at the
human--computer interface: whether a user can tell that an output is
trustworthy, whether sources can be verified, whether confident presentation
masks error, and whether the user can steer the system back toward accuracy.
The empirical literature on human--AI decision-making shows that these are not
secondary concerns. Users over-rely on confidently phrased but incorrect model
output \citep{Si:2024:Verify}, adjust reliance in response to expressed
uncertainty \citep{Kim:2024:Uncertainty}, and treat surface features such as
lexical alignment as evidence of faithfulness \citep{Chiesurin:2023:Dangers}.
A system's measured factuality and its \emph{perceived} reliability are
therefore distinct quantities, and only the first is currently measured in a
standardized way. This is precisely the gap that human-centered AI research has
identified as decisive for adoption and safe use: well-designed systems must be
not only capable but \emph{reliable, safe, and trustworthy} from the user's
standpoint \citep{Shneiderman:2020:HCAI}.

Human-centered evaluation has long addressed analogous challenges with
lightweight, standardized instruments that turn subjective experience into
systematically interpretable measurements. The System Usability Scale (SUS)
\citep{Brooke:1996:SUS,Lewis:2018:SUS,Bangor:2008:Empirical} made perceived
usability comparable across systems with only ten items, and the System
Causability Scale (SCS) \citep{Holzinger:2020:Causability} extended the same
philosophy to the perceived quality of explanations in human--AI interaction
\citep{Holzinger:2019:Causability,Plass:2023:Pathology}. No comparable
instrument yet exists for the rapid, structured assessment
of \emph{perceived reliability}---how hallucination manifests from the user's
perspective.

We address this gap with the \textbf{System Hallucination Scale (SHS)}, a
ten-item, five-point Likert instrument for assessing perceived reliability of
LLM output across five interpretable dimensions. Following SUS and SCS, SHS is
deliberately compact, domain-agnostic, and interpretable without training.

\subsection{Contributions}

Our contribution is threefold, and it is deliberately narrow. The claim we
make is that a ten-item, five-point instrument in the SUS/SCS mould can be
applied to the perceived reliability of LLM output and will behave the way a
well-formed scale of that family behaves. Establishing that---and nothing
larger---is the purpose of this paper.

\begin{enumerate}[leftmargin=*,itemsep=2pt]
\item \textbf{An instrument.} We define a ten-item scale whose five dimensions
are derived from documented hallucination failure modes rather than from a
generic reliability construct, and we ground it in the psychometric tradition
of human-centered evaluation (Sections~3 and~4).

\item \textbf{A scoring procedure.} We give a precise, reproducible scoring
algorithm yielding a directional score on $[-1,+1]$ together with a
paired-item consistency diagnostic, released with an open reference
implementation and a multilingual interactive calculator (Section~5).

\item \textbf{A proof of concept, with its boundaries stated.} We report a
supervised validation study ($N = 193$ participants, 386 LLM evaluations,
$n = 47$ trained administrators) that establishes content, internal-structure,
and response-process evidence for the instrument, and we specify explicitly
which forms of validity evidence remain outstanding and how they should be
obtained (Sections~6--8). The study is a demonstration that the items work, in
the same sense in which the original SUS and SCS papers demonstrated that their
items worked \citep{Brooke:1996:SUS,Holzinger:2020:Causability}; it is not a
study of hallucination.
\end{enumerate}

\subsection{What This Paper Does Not Claim}
\label{sec:nonclaims}

A short instrument invites large expectations. In the interest of scoping
claims to evidence, we state six non-claims explicitly, and we hold to them
throughout.

\begin{enumerate}[label=(N\arabic*),leftmargin=*,itemsep=2pt]
\item \textbf{SHS is not a hallucination detector.} It records human judgment;
it does not identify, localize, or count hallucinated content.

\item \textbf{This is not a hallucination study.} We measure how reliable LLM
output was judged to be, not how reliable it was. No ground truth was
established for any response, and no hallucination rate is estimated anywhere
in this paper.

\item \textbf{We do not benchmark models.} The study was designed to validate
an instrument, not to rank systems; the five systems are documented
(Section~\ref{sec:systems}) for replication, not for comparison, and all
analyses pool across them.

\item \textbf{This is neither a naturalistic field study nor a controlled
laboratory experiment.} Participants worked under a supervised, adversarially
oriented protocol, which rules out the first; they composed their own prompts,
so the stimulus was not held constant, which rules out the second
(Section~\ref{sec:design}).

\item \textbf{We do not claim criterion validity.} SHS scores have not been
correlated with any objective hallucination measure. Whether perceived
reliability tracks measured reliability is an open empirical question, and
Section~\ref{sec:criterion} specifies the study that would answer it.

\item \textbf{We do not claim population generality.} The sample is a
convenience sample of German- or English-speaking adults without professional
AI/ML experience, recruited through university and online channels
(Section~\ref{sec:participants}).
\end{enumerate}

What remains after these six subtractions is the claim we do make: ten short
items, scored in a fixed way, are understood by ordinary users, are completed
in about four minutes, and yield internally consistent and structurally
coherent measurements. That is the same claim SUS made in 1996, and it is
sufficient to make an instrument useful.

% =====================================================================
\section{Hallucination in Large Language Models: A Critical Review}
\label{sec:background}
% =====================================================================

This section reviews the state of the art on LLM hallucination and, for each
strand, states what it establishes and where it falls short. The argument we
build is cumulative: the technical literature has produced increasingly precise
accounts of \emph{what} hallucination is and \emph{how often} it occurs, while
leaving largely unmeasured the question that determines its practical
consequences---what a user makes of it.

\subsection{From a Clinical Metaphor to a Contested Construct}

The term \emph{hallucination} originates in psychiatry, where it denotes vivid
perceptual experiences without an external stimulus
\citep{AmericanPsychiatricAssociation:2013:DSM5,Haddock:1999:PSYRATS}.
Clinically, hallucinations are assessed through structured instruments
capturing modality, intensity, frequency, and subjective impact
\citep{Waters:2017:Hallucinations}. We invoke this lineage only
instrumentally---it demonstrates that complex, subjective phenomena can be
\emph{operationalized} through standardized scales---without implying any
cognitive or phenomenological equivalence between humans and machines.

The borrowing is, however, contested, and the contest is substantive rather
than merely terminological. \citet{Maleki:2024:Misnomer} survey definitions
across fourteen bibliographic databases and find no consistent usage.
\citet{Venkit:2024:Audit} audit 103 NLP publications and survey 171
practitioners, likewise documenting definitional drift and, notably, a gap
between how researchers define the term and how affected users experience the
phenomenon. \citet{Hicks:2024:Bullshit} argue on philosophical grounds that
``hallucination'' mischaracterizes the failure, since the model is indifferent
to truth rather than misperceiving. We take these critiques seriously: they are
the reason Section~\ref{sec:construct} fixes a single construct term and
separates the \emph{phenomenon} being probed from the \emph{quantity} being
measured. They do not, however, dissolve the measurement problem. Whatever the
phenomenon is called, users encounter confident falsehoods and must decide what
to do about them.

\subsection{Mechanisms: Why Language Models Hallucinate}

Three complementary explanations have emerged, and they have different
implications for what mitigation can achieve.

The \emph{statistical-objective} account holds that hallucination is a
predictable consequence of how models are trained and scored.
\citet{Kalai:2025:WhyHallucinate} show that pretraining error rates on
next-token prediction lower-bound the rate of confident falsehoods, and---more
consequentially for evaluation practice---that binary, accuracy-style scoring
actively rewards guessing over abstention. On this account hallucination is
partly an artifact of the benchmark culture that measures it.

The \emph{computability} account is stronger still.
\citet{Xu:2024:Inevitable} formalize hallucination as inconsistency between a
computable model and a computable ground-truth function and prove that no LLM
can be free of it across all inputs. If correct, the practical target is
management and disclosure, not elimination.

The \emph{alignment} account locates a further source in post-training.
Reinforcement learning from human feedback optimizes for judgments that humans
reward \citep{Bai:2022:Training}, and human raters demonstrably prefer
responses that agree with them; \citet{Sharma:2024:Sycophancy} document
systematic sycophancy across five deployed assistants. This matters directly
for the present work: it predicts that a model's apparent \emph{responsiveness
to corrective guidance}---a dimension of SHS---may reflect accommodation rather
than correction, which is exactly why the dimension must be measured rather
than assumed.

Against these pessimistic results stands evidence that models carry usable
internal signals of their own uncertainty: larger models are reasonably
calibrated on multiple-choice and true/false items and can self-evaluate
\citep{Kadavath:2022:KnowWhatTheyKnow}. Calibration, however, is a property of
the model's internal probabilities and is not, by itself, communicated to the
user.

\subsection{Taxonomies and Their Limits}

In natural language processing, hallucination first became a concern for neural
sequence-to-sequence models in summarization and translation
\citep{Maynez:2020:Faithfulness}, and became acute with large autoregressive
transformers \citep{Huang:2025:Hallucination}. Surveys distinguish
\emph{intrinsic} hallucinations, arising from internal inconsistency, from
\emph{extrinsic} hallucinations, involving unsupported external facts
\citep{Ji:2023:HallucinationSurvey,Zhang:2025:SirensSong}, and a parallel
distinction separates \emph{faithfulness} to a provided source from
\emph{factuality} with respect to world knowledge.
\citet{Rawte:2023:Troubling} extend this into a graded taxonomy with degrees of
severity.

These taxonomies are indispensable for annotation, but they share a structural
property that limits their reach: each category is defined by the relationship
between the output and a reference. They classify the \emph{text}. They say
nothing about the relationship between the output and the reader, and it is
that relationship---not the annotation label---that determines whether a
hallucination causes harm. A subtle, well-hedged extrinsic hallucination that a
user notices and discounts is a different event from a confidently asserted one
that the user acts upon, although both receive the same label.

\subsection{Detection: What Works, and Where It Falls Short}

Detection approaches fall into four families.

\emph{Reference-based verification} checks claims against evidence. FEVER
\citep{Thorne:2018:FEVER} verifies claims against Wikipedia; FActScore
\citep{Min:2023:FActScore} decomposes generations into atomic facts and scores
each against a knowledge base. These give precise, auditable judgments---but
only where a suitable reference corpus exists, which excludes most open-ended
advice, reasoning, and forecasting.

\emph{Benchmark suites} standardize the inputs. TruthfulQA
\citep{Lin:2022:TruthfulQA} probes imitative falsehoods, HaluEval
\citep{Li:2023:HaluEval} supplies large-scale binary labels, SimpleQA
\citep{Wei:2024:SimpleQA} targets short-form factuality, and RAGTruth
\citep{Niu:2024:RAGTruth} annotates hallucinations in retrieval-augmented
output at the span level. Their weakness is the converse of their strength:
static prompt sets are, by construction, not the prompts any particular user
will issue, and performance on them can be optimized directly.

\emph{Reference-free methods} avoid the ground-truth requirement.
SelfCheckGPT \citep{Manakul:2023:SelfCheckGPT} exploits inconsistency across
stochastic samples; semantic entropy \citep{Farquhar:2024:SemanticEntropy}
computes uncertainty over meaning-equivalence clusters rather than token
sequences and detects confabulations across tasks without task-specific
supervision; related work maps formal uncertainty in reasoning tasks
\citep{Ganguly:2025:Grammars}. These are the most deployable technical
detectors available, but they detect \emph{inconsistency}, which is a proxy for
error: a model that is consistently wrong is invisible to them.

\emph{Model-as-judge} approaches \citep{Liu:2023:GEval}, and trained
faithfulness classifiers such as HHEM \citep{Hughes:2024:HHEM} or the
retrieval-oriented RAGAS metrics \citep{Es:2024:RAGAS}, scale human-like
judgment cheaply. They inherit the failure modes of the judging model and
require human calibration to be trusted---using a system that hallucinates to
adjudicate hallucination.

Two properties are common to all four families. First, they assess
system-internal correctness on prompt--response pairs, typically offline.
Second, they report one or two dimensions. Neither property is a defect for
their intended purpose; both mean that a different instrument is needed for a
different question.

\subsection{Mitigation: Reduction, Not Elimination}

Retrieval-augmented generation grounds output in retrieved documents
\citep{Lewis:2020:RAG} and measurably reduces hallucination in conversation
\citep{Shuster:2021:Retrieval}; the mitigation literature now spans more than
thirty distinct techniques \citep{Tonmoy:2024:Mitigation}. The evidence
consistently shows reduction rather than removal---RAGTruth exists precisely
because retrieval-augmented systems still hallucinate
\citep{Niu:2024:RAGTruth}---and the theoretical results above suggest that this
is not a temporary state of the engineering art. Mitigation also introduces its
own interaction problem: a system that cites retrieved sources may raise
perceived reliability more than it raises actual reliability, since a citation
is easier to display than to verify.

\subsection{The User in the Loop: What Remains Unmeasured}

Human-subject research on AI-assisted decision-making establishes three
findings that bear directly on hallucination.

First, \emph{reliance is not calibrated by accuracy alone}. Users over-rely on
AI output, and interventions that raise verification effort---cognitive forcing
functions \citep{Bucinca:2021:CognitiveForcing}, explanations that genuinely
lower verification cost \citep{Vasconcelos:2023:Overreliance}---can reduce
over-reliance, whereas explanations that merely look authoritative do not.
Second, \emph{presentation governs trust}. Expressed uncertainty shifts
reliance and trust \citep{Kim:2024:Uncertainty}; superficial cues such as
lexical alignment inflate perceived faithfulness
\citep{Chiesurin:2023:Dangers}; and LLM assistance improves human verification
of claims \emph{except} when the model is convincingly wrong, in which case it
degrades it \citep{Si:2024:Verify}. Third, \emph{trust is a distinct construct
from correctness} with its own determinants and its own measurement tradition
\citep{Lee:2004:TrustAutomation,Jacovi:2021:FormalizingTrust}, and
transparency research argues for human-centered measurement of how model
behavior is understood in context \citep{Liao:2024:Transparency}.

Subjective human judgment is not a novelty in this literature; it is its
foundation. Human ratings have anchored evaluation in HCI and NLP for decades,
from conversational quality \citep{See:2019:GoodConversation} to summarization
\citep{Narayan:2018:XSum} and translation \citep{Koehn:2004:Significance}, and
expert evaluation with curated prompt--response sets adds rigor in high-stakes
domains such as medicine and law
\citep{Weng:2017:Clinical,Liu:2025:Diagnosis,Zhong:2020:Legal}. Their strength
is that they capture trust and the lived quality of interaction rather than
abstract correctness alone; their weaknesses are cost, limited scalability, and
susceptibility to rater bias, restricted domain knowledge, and framing effects.
What is missing from the hallucination literature is therefore not human
judgment but a \emph{standardized and compact} form of it---precisely the gap
that SUS closed for usability. It is also worth recording that
hallucination-like patterns are not unique to machines: internal incoherence,
unverifiability, and implausible reasoning characterize misinformation
\citep{Vosoughi:2018:Spread}, polarized discourse \citep{Bail:2018:Exposure},
and disordered thought \citep{Corcoran:2020:Disorders}, and the professional
fact-checking infrastructure built to counter them is itself being
reconsidered in light of LLMs \citep{Augenstein:2024:Factuality}. Reliability,
in the end, is a judgment made by a situated human reader.

Evidence from deployment points the same way: analyses of user-reported errors
``in the wild'' in app-store reviews show that users articulate hallucination
complaints in structured, recurring categories---fabricated content, missing or
invented sources, confident wrongness, failure to correct on request
\citep{Massenon:2025:UserReported}. Users are already making these judgments.
What is missing is a standardized way to record them.

Taken together, these strands identify a measurable quantity that no existing
instrument captures in a comparable, repeatable form: the reliability of an
LLM's output \emph{as judged by the person reading it}, resolved into the
distinct failure modes that the technical literature has catalogued. SHS is
designed to fill exactly this role, inheriting the simplicity and
interpretability that made SUS a standard
\citep{Brooke:1996:SUS,Lewis:2018:SUS} and aligning with the broader agenda of
designing AI that is trustworthy and accessible to a diverse user population
\citep{Shneiderman:2020:HCAI,Stephanidis:1999:InformationSociety}.

% =====================================================================
\section{Construct, Terminology, and Scope}
\label{sec:construct}
% =====================================================================

\subsection{One Construct, One Term}
\label{sec:terminology}

Throughout this paper the construct that SHS measures is called
\textbf{perceived reliability}, defined as:

\begin{quote}
\emph{the degree to which a human reader, after interacting with an LLM, judges
its output to be factually correct, traceable to verifiable sources, logically
coherent, transparently rather than deceptively presented, and correctable
through further instruction.}
\end{quote}

We use this term, and only this term, for the measured quantity. Three
distinctions are held fixed for the remainder of the paper:

\begin{itemize}[leftmargin=*,itemsep=2pt]
\item \textbf{Hallucination} denotes the \emph{phenomenon}: model output that
is factually incorrect, weakly grounded, internally incoherent, or misleading
while remaining superficially fluent. It is a property of the system's output.
\item \textbf{Reliability} denotes an \emph{objective} property
of a system, quantified by automated benchmarks. This paper measures no such
quantity.
\item \textbf{Perceived reliability} denotes the \emph{judgment} a user forms
about that output. It is what SHS measures.
\end{itemize}

A higher SHS score means higher perceived reliability. We deliberately avoid
the phrase ``hallucination risk'' as a name for the measured quantity, since it
implies a probability of an objective event that SHS does not estimate. 

\subsection{Why the Instrument Is Called a Hallucination Scale}
\label{sec:naming}

The instrument is named for the phenomenon it interrogates, not for the
epistemic status of the number it produces. This is the established convention
in the family of scales SHS belongs to, and the objection that the name
overpromises applies to that family as a whole or to none of it. The System
Usability Scale does not measure usability in the objective sense of task
completion time or error rate; it measures \emph{perceived} usability
\citep{Brooke:1996:SUS,Bangor:2008:Empirical}. The System Causability Scale
does not measure whether an explanation is causally correct; it measures
\emph{perceived} explanation quality \citep{Holzinger:2020:Causability}. The
NASA Task Load Index does not measure workload physiologically. In each case
the scale name identifies the target construct domain, and the word
``perceived'' is carried by the instrument's definition and reporting
conventions rather than by its title.

There is, however, a stronger and more specific argument, and we consider it
the decisive one: \emph{the item structure is hallucination-specific, not
generic}. A general-purpose perceived-reliability scale would ask about
correctness and consistency. SHS instead asks whether sources were omitted or
invented (Q4), whether falsehoods were presented confidently and misleadingly
(Q8), and whether the system persisted in error after corrective instruction
(Q10). These are the documented signatures of hallucination specifically:
source fabrication, fluent confident wrongness, and resistance to correction.
They correspond one-to-one with the failure modes catalogued in the surveys
reviewed in Section~\ref{sec:background} and with the categories users
spontaneously produce when reporting hallucination in deployment
\citep{Massenon:2025:UserReported}. An instrument whose items are derived from
hallucination phenomenology is properly named after hallucination, in the same
way that a depression inventory is named after depression although it measures
self-reported symptoms rather than neurochemistry.

We considered two alternatives. ``Perceived LLM Reliability Scale'' is more
literal but loses the connection to the specific failure phenomenology and
would suggest a broader construct than the items support. ``Perceived
Hallucination Scale'' is accurate but breaks the naming symmetry with SUS and
SCS, whose joint administration we suggest (Section~\ref{sec:family}). We therefore retain
\emph{System Hallucination Scale} and discharge the ambiguity by definition:
the abstract, Section~\ref{sec:terminology}, and every reported result state
that the measured quantity is perceived reliability.

\subsection{Scope}

SHS applies where a human has read LLM output and can form a judgment about it.
It does not require ground truth, a reference corpus, or model access, which is
what makes it deployable in settings---clinical consultation, legal drafting,
educational use---where no benchmark exists
\citep{Rajpurkar:2022:AIHealth,Mueller:2021:TenCommandments}. It is
correspondingly limited by the rater's knowledge: a user who cannot detect an
error cannot report it, and SHS will record high perceived reliability for a
system that is confidently and undetectably wrong. This is not a defect to be
engineered away but the definition of the construct, and it is the reason SHS
must be read alongside, never instead of, automated measurement.

% =====================================================================
\section{The System Hallucination Scale (SHS)}
% =====================================================================

The scale comprises ten items organized into five conceptual dimensions:
factual accuracy, source reliability, logical coherence, deceptiveness of
presentation, and responsiveness to user guidance. Each dimension is
represented by one positively and one negatively worded item. This paired
structure, adopted from SUS and SCS, serves two purposes: reducing acquiescence
bias and enabling internal consistency diagnostics
\citep{Boateng:2018:BestPractices}. Each item is rated on a five-point Likert
scale. The complete instrument is given in Table~\ref{tab:shs_items}.

\begin{table}[htbp]
\centering
\caption{The ten items of the System Hallucination Scale (SHS), grouped by
dimension. Each dimension is measured by one positively worded ($+$) and one
negatively worded ($-$) item; this bipolar pairing reduces acquiescence bias
and supports the consistency diagnostic
\citep{Brooke:1996:SUS,Lewis:2018:SUS}.}
\label{tab:shs_items}
\small
\renewcommand{\arraystretch}{1.25}
\begin{tabular}{@{}>{\raggedright\arraybackslash}p{2.7cm}c>{\raggedright\arraybackslash}p{9.4cm}@{}}
\toprule
\textbf{Dimension} & \textbf{Pol.} & \textbf{Item} \\
\midrule
\multirow{2}{=}{Factual Accuracy}
 & $+$ & Q1. The response was factually reliable. \\
 & $-$ & Q2. The LLM frequently generated false or fabricated information. \\
\addlinespace[2pt]
\multirow{2}{=}{Source Reliability}
 & $+$ & Q3. It was easy to find and verify the sources of the presented information. \\
 & $-$ & Q4. The LLM often omitted sources or invented them, and it was difficult to recognize what was real. \\
\addlinespace[2pt]
\multirow{2}{=}{Logical Coherence}
 & $+$ & Q5. The LLM's reasoning was logically structured and supported by facts. \\
 & $-$ & Q6. The LLM's reasoning contained unfounded or illogical steps. \\
\addlinespace[2pt]
\multirow{2}{=}{Deceptiveness}
 & $+$ & Q7. False or fabricated information was easy to recognize. \\
 & $-$ & Q8. The LLM presented false information in a confident and misleading manner. \\
\addlinespace[2pt]
\multirow{2}{=}{Responsiveness to Guidance}
 & $+$ & Q9. I was able to prompt the LLM to provide more accurate answers when needed. \\
 & $-$ & Q10. The LLM ignored my instructions and continued to generate false information. \\
\bottomrule
\end{tabular}
\end{table}

Each dimension corresponds to a failure mode documented in
Section~\ref{sec:background}. \emph{Factual accuracy} captures correctness and
freedom from fabrication---the extrinsic hallucination of the NLP taxonomies
\citep{Ji:2023:HallucinationSurvey}. \emph{Source reliability} addresses
traceability and verifiability, the property that reference-based verification
\citep{Thorne:2018:FEVER,Min:2023:FActScore} operationalizes technically and
that users report as ``invented citations'' in deployment
\citep{Massenon:2025:UserReported}. \emph{Logical coherence} concerns the
internal structure of reasoning---the intrinsic hallucination of the
taxonomies---distinguishing fluent text from defensible argument.
\emph{Deceptiveness} captures the interaction-critical property that the
technical literature does not measure at all: whether errors announce
themselves or are asserted with unwarranted confidence, which is the condition
under which LLM assistance actively degrades human verification
\citep{Si:2024:Verify,Chiesurin:2023:Dangers}. \emph{Responsiveness to
guidance} reflects controllability in human-in-the-loop settings, and is
motivated directly by the sycophancy findings of
\citet{Sharma:2024:Sycophancy}: whether corrective prompting produces genuine
correction is an empirical question about a deployed system, and a user is
positioned to observe it.

% =====================================================================
\section{Scoring Methodology and Algorithm}
% =====================================================================

\subsection{Input Encoding and Dimension Structure}

Each item is rated on a five-point Likert scale and encoded symmetrically
as the integers $\{-2,-1,0,+1,+2\}$, ranging from strongly disagree to
strongly agree. Positively and negatively worded items are paired by design.
The ten items map to five dimensions: Factual Accuracy (Q1, Q2), Source
Reliability (Q3, Q4), Logical Coherence (Q5, Q6), Deceptiveness (Q7, Q8), and
Responsiveness to Guidance (Q9, Q10).

\subsection{Scoring Logic}

Let $p_i$ denote the response to the positive item and $n_i$ the response to the
negative item of dimension~$i$. The \emph{dimension score} is the normalized
difference
\[
s_i = \frac{p_i - n_i}{4} \in [-1, +1],
\]
where higher values indicate higher perceived reliability. A \emph{consistency
indicator} is computed as
\[
c_i = \frac{p_i + n_i}{4}.
\]
Values of $c_i$ near zero indicate internally coherent paired judgments; larger
absolute values reflect ambiguity or mixed impressions. Consistency indicators
are diagnostic only and are not part of the aggregate score. The overall SHS
score is the mean of the five dimension scores,
\[
\mathrm{SHS} = \frac{1}{5}\sum_{i=1}^{5} s_i \in [-1, +1],
\]
which can optionally be rescaled for comparability with SUS-style ranges via
$\mathrm{SHS}_{100} = 50 \times (\mathrm{SHS} + 1)$.

\subsection{Interpretation}

Table~\ref{tab:interpretation} gives interpretation bands, stated consistently
in terms of perceived reliability (Section~\ref{sec:terminology}). Consistency
indicators with $|c_i| > 0.25$ may signal rater uncertainty, item
misunderstanding, or context-dependent judgments; by separating directional
scoring from consistency, SHS functions both as a measurement and as a
quality-control tool for human ratings. A complete Python reference
implementation, an interactive multilingual calculator, and all materials are
openly available (Supplementary Material~S1).

We emphasize one interpretive constraint. Because the bands describe perceived
rather than measured reliability, a score in the highest band licenses the
statement ``users judged this system's output to be reliable under these
conditions'', not ``this system is reliable''. Absolute-band language should be
used only in comparative or longitudinal settings, where the rater population
and prompting regime are held constant.

\begin{table}[htbp]
\centering
\caption{Interpretation guidelines for SHS scores. Bands describe
\emph{perceived} reliability under the prompting regime and rater population in
which the score was obtained; they are not statements about objective model
reliability.}
\label{tab:interpretation}
\begin{tabular}{ccp{7.6cm}}
\toprule
\textbf{SHS} & \textbf{SHS$_{100}$} & \textbf{Interpretation} \\
\midrule
$[+0.5, +1.0]$ & $[75, 100]$ & High perceived reliability \\
$[+0.0, +0.5)$ & $[50, 75)$ & Moderate perceived reliability; some concerns \\
$[-0.5, +0.0)$ & $[25, 50)$ & Low perceived reliability; caution advised \\
$[-1.0, -0.5)$ & $[0, 25)$ & Very low perceived reliability \\
\bottomrule
\end{tabular}
\end{table}

% =====================================================================
\section{Method: Supervised Validation Study}
\label{sec:method}
% =====================================================================

\subsection{Design and Its Rationale}
\label{sec:design}

The study assessed the feasibility, clarity, and psychometric robustness of SHS
as a measurement instrument. We state plainly what it is not: it is
\emph{not} an objective hallucination-detection study (it measures perceived,
not ground-truth, reliability); it is \emph{not} a model benchmark (systems
are pooled, not compared); it is \emph{not} a naturalistic field deployment
(prompts were solicited under an adversarial protocol); and it is \emph{not} a
controlled laboratory experiment in the strict sense (stimulus wording was not
held constant across participants). We describe it as
a \textbf{supervised validation study}: an administrator-guided protocol in
which participants interacted with a deployed LLM under a structured,
two-category prompt protocol (Section~\ref{sec:protocol}) deliberately
designed to elicit hallucination-prone behavior.

This design was chosen for a specific psychometric reason, and its costs are
real. Estimating internal consistency and inter-dimension structure requires
variance on every dimension. A naturalistic protocol, in which most
interactions are unproblematic, would produce ceiling-heavy distributions on
all five dimensions and would make the reliability coefficients
uninformative---the classic restriction-of-range problem in scale development
\citep{Boateng:2018:BestPractices}. Prompt content varied \emph{across} the
193 participants' own sessions (Section~\ref{sec:protocol}), which is a
source of stimulus heterogeneity we address directly in
Section~\ref{sec:limitations}.

The cost is ecological validity, and we state it plainly rather than describing
the protocol as ``realistic''. Section~\ref{sec:ecological} sets out exactly
what the design does and does not license.

\subsection{Systems Evaluated}
\label{sec:systems}

Each participant was randomly assigned two of five widely used commercial
systems, interacted with each in turn, and completed a full SHS
administration immediately after each interaction (Section~\ref{sec:protocol}).
Table~\ref{tab:systems} reports the resulting per-system evaluation counts
after exclusion of coordinator test-run entries (see
Section~\ref{sec:participants}).

\begin{table}[htbp]
\centering
\caption{Systems evaluated in the supervised validation study. Each of
$N = 193$ participants evaluated two of the five systems (random assignment),
yielding 386 SHS evaluations in total.}
\label{tab:systems}
\small
\begin{tabular}{@{}>{\raggedright\arraybackslash}p{4.6cm}>{\raggedright\arraybackslash}p{3.4cm}>{\raggedright\arraybackslash}p{3.0cm}c@{}}
\toprule
\textbf{System (model, version)} & \textbf{Access mode} & \textbf{Collection window} & \textbf{Evaluations} \\
\midrule
ChatGPT (OpenAI, {GPT-4}-based) & Free-tier web interface & Oct--Nov 2025 & 68 \\
Claude (Anthropic, Claude 3 Sonnet) & Free-tier web interface (Claude.ai) & Oct--Nov 2025 & 85 \\
Gemini (Google) & Free-tier web interface & Oct--Nov 2025 & 78 \\
Mistral (Mistral AI) & Public inference platform & Oct--Nov 2025 & 79 \\
DeepSeek (DeepSeek AI) & Public inference platform & Oct--Nov 2025 & 76 \\
\midrule
\textbf{Total} & & & \textbf{386} \\
\bottomrule
\end{tabular}
\end{table}

All systems were accessed in their free or publicly available consumer
version \citep[for the underlying model of the first, see][]{OpenAI:2023:GPT4}
rather than via API with standardized parameters, so as to capture
user perception of LLMs as encountered in ordinary practice (default
settings, interface design, typical interaction patterns) rather than under
controlled generation parameters. Model identity is reported here for
replication, not for comparison. Because the validation analyses in
Section~\ref{sec:results} pool across systems, any between-system differences
in hallucination behavior enter those analyses as additional stimulus
variance. We treat this explicitly in Section~\ref{sec:limitations}: pooling
inflates the observed spread of dimension scores and is therefore
conservative with respect to internal consistency, but it means the reported
inter-dimension correlations describe the instrument's behavior across a
heterogeneous stimulus set rather than for any single system. Model-resolved
comparisons are the subject of a companion analysis and are deliberately not
reported here.

\subsection{Interaction Protocol and Prompt Design}
\label{sec:protocol}

Each session opened with an introduction and informed consent
(5~min). The participant then completed two interaction blocks, one per
assigned system (Section~\ref{sec:systems}), each consisting of an LLM
interaction (15~min) followed immediately by a full SHS administration
(5~min). The session closed with the administrator feedback questionnaire on
the instrument (5~min) and demographic items (2~min). The two blocks are the
reason the 193 participants yield 386 evaluations.

Rather than a fixed, researcher-scripted list of prompts, each participant
composed their own prompts during the session under experimenter guidance,
following a two-category protocol applied consistently across all
administrators (Table~\ref{tab:prompts}). For each LLM interaction,
participants posed a minimum of three \emph{Verified Knowledge} prompts,
drawn from topics they knew or could easily check---local or regional facts,
professional or domain expertise, or easily verifiable historical or
scientific facts---and a minimum of three \emph{``Glatteis''} (German for
``black ice'', i.e.\ a trick or trap) prompts, deliberately constructed to
reference non-existent people or entities, impossible events, fabricated
facts, or category errors, so that the ideal response is to flag the false
premise rather than hallucinate an answer. Administrators provided worked
examples for both categories and helped participants formulate suitable
prompts when needed, but did not prescribe the wording. Participants were
additionally encouraged to probe with follow-up questions, explicit requests
for sources, and challenges to answers they judged suspicious. This
participant-generated design was a deliberate trade-off: it sacrifices exact
stimulus control in order to obtain stimulus diversity and, more importantly,
to let participants draw on their own domain expertise, where a subtle
fabrication is detectable and a generic pre-written question would not reach.
It does not thereby buy ecological validity---the session remains supervised
and adversarially oriented (Section~\ref{sec:ecological}).

\begin{table}[htbp]
\centering
\caption{Structure of the two-category prompt protocol used in every LLM
interaction, and the SHS dimensions each category primarily targets.
Participants composed their own items within each category (minimum three per
category per interaction). Illustrative items are participants' verbatim
prompts, translated from German; the complete item set (German originals and
English translations), together with the instructions and worked examples given
to administrators, is deposited in the project repository. FA = Factual Accuracy, SR = Source Reliability, LC =
Logical Coherence, DP = Deceptiveness, RG = Responsiveness to Guidance.}
\label{tab:prompts}
\small
\renewcommand{\arraystretch}{1.2}
\begin{tabular}{@{}>{\raggedright\arraybackslash}p{3.2cm}>{\raggedright\arraybackslash}p{7.1cm}>{\raggedright\arraybackslash}p{2.2cm}@{}}
\toprule
\textbf{Category} & \textbf{Purpose} & \textbf{Dimensions} \\
\midrule
Verified Knowledge (``Wissen'') & Objectively checkable prompts on topics the participant knew or could easily verify, establishing a correctness baseline the participant could adjudicate. & FA, SR \\
``Glatteis'' (Trick) & Prompts referencing non-existent people or entities, impossible events, fabricated facts, or category errors, constructed so that hallucinating an answer (rather than flagging the false premise) indicates problematic behavior. & DP, LC, RG \\
\bottomrule
\end{tabular}
\end{table}

\noindent\emph{Illustrative items} (participants' verbatim prompts, translated
from German):

\begin{itemize}[nosep]
\item \textbf{Verified Knowledge:} ``What does the Austrian Forest Act
regulate with respect to protective forests?''; ``What is the area of
Austria?''
\item \textbf{``Glatteis'':} ``The Earth is a cube.''; ``Describe in detail
the scientific career of the famous fictional chemist Dr.\ Alistair Finch
from the novel `Der chromatische Algorithmus.' Focus on his groundbreaking
`Finch constant' describing the subatomic chirality of vacuum polarons, and
give three quotations from peer-reviewed articles from 2014--2017 that
criticize his work.''
\end{itemize}

\subsection{Participants and Recruited Population}
\label{sec:participants}

A total of $N = 193$ participants completed the protocol, administered by
$n = 47$ trained student experimenters (coordinator self-test entries used to
pilot the survey instrument prior to recruiting real participants were
excluded from this count). Each participant was randomly assigned two of the
five systems in Table~\ref{tab:systems} and completed a full SHS
administration after each interaction, yielding 386 evaluations in total. This
sample comfortably exceeds the size required for stable estimates of
questionnaire reliability and correlational structure in applied user research
\citep{Sauro:2016:Quantifying}.

Participants were recruited through university announcements and online
advertisements. Inclusion criteria were age 18 or older, fluency in German or
English, and no prior professional experience in AI/ML development; exclusion
criteria were participation in prior SHS validation studies and employment at
companies developing LLM products (Supplementary Material~S5).

We characterize this sample precisely rather than describing it as diverse. It
is a convenience sample of German- or English-speaking adults recruited through
university and online channels and screened to exclude AI/ML professionals.
This composition serves the study's purpose---it approximates the non-expert
end-user for whom the instrument is intended, and the exclusion of AI
professionals removes a plausible source of expert-specific interpretation of
the items---but it does not support claims about heterogeneity of age,
education, cultural background, or domain expertise, and no such claims are
made in this paper. Whether SHS behaves equivalently across demographic
subgroups is an open question requiring measurement-invariance testing on a
stratified sample (Section~\ref{sec:limitations}).

The study was approved by the Ethics Committee of the Medical University of
Graz (EK-Number 34-527 ex 21/22); all participants gave informed consent and
data were collected anonymously.

\subsection{Rating Procedure}
\label{sec:rating}

Each LLM interaction was rated once, immediately after the interaction, by the
participant who conducted it. The unit of analysis is therefore the
\emph{evaluation}: one participant, one system, one SHS administration, giving
386 evaluations from 193 participants.

This single-rating design follows from the construct. SHS measures the
perceived reliability of output \emph{as judged by the person who read it}
(Section~\ref{sec:terminology}). That judgment is not an estimate of an
external quantity on which independent observers could converge, and a second
rater cannot supply it: an observer watching someone else's session can report
their own impression, but not that participant's. Inter-rater agreement is
consequently not a meaningful property of this instrument, for the same reason
that SUS and SCS report internal consistency but no inter-rater coefficient
\citep{Brooke:1996:SUS,Lewis:2018:SUS,Holzinger:2020:Causability}. The
appropriate reliability evidence for a self-report scale of this family is
internal consistency and internal structure, both reported in
Section~\ref{sec:results}, and---in future work---test--retest stability
(Section~\ref{sec:outstanding}).

\subsection{Analysis Plan}

We evaluated the instrument's psychometric quality following established
practice for standardized questionnaires in human--computer interaction
\citep{Sauro:2016:Quantifying,Cairns:2019:Statistics} and the
reliability-reporting conventions used for usability scales such as SUS
\citep{Lewis:2018:SUS}. We organize the evidence using the standard validity
framework \citep{Cronbach:1955:ConstructValidity,Messick:1995:Validity}:
content evidence, internal-structure evidence, response-process evidence, and
criterion evidence. All tests were two-sided with
$\alpha = .05$, and dimension scores were treated as interval quantities on the
symmetric $[-1,+1]$ scale defined in Section~5. Reliability was interpreted
against the conventional benchmarks ($\geq .70$ acceptable, $\geq .80$ good)
recommended for applied HCI measurement \citep{Sauro:2016:Quantifying}.

\subsection{Ecological Validity: What This Design Does and Does Not Establish}
\label{sec:ecological}

Because the protocol is supervised and adversarially oriented, and because
its items were composed by participants rather than fixed in advance, we set out
its evidentiary reach explicitly.

\paragraph{What the design supports.} Estimates of item clarity,
administration time, internal consistency, inter-dimension structure, and
response-process quality \emph{for this instrument}, obtained under
conditions in which all five dimensions could vary. It supports the
conclusion that administrators report the items to be understood as intended.

\paragraph{What the design does not support.} Three things. (i) Estimates of
how reliable LLMs appear in ordinary use: the protocol is a stress test,
so the absolute score level reported in Section~\ref{sec:profile} is a
lower bound conditional on adversarial prompting, not a population estimate.
(ii) Claims about fully self-directed interaction: although participants
composed their own prompts (Section~\ref{sec:protocol}), they did so under an
experimenter-guided, two-category protocol constructed to elicit
hallucination-prone behavior, so the study says less about behavior when users
pursue open-ended goals of their own choosing, and the presence of an
administrator may itself have raised scrutiny.
(iii) Any inference from perceived to objective reliability
(Section~\ref{sec:criterion}).

\paragraph{Consequences for the reported statistics.} Adversarial prompting
raises the observed variance of dimension scores, which mechanically favors
higher internal-consistency estimates than a naturalistic protocol would
yield. The coefficients reported below should therefore be read as
obtained under favorable variance conditions, and re-estimation under
naturalistic use is a required next step, not an optional extension.

% =====================================================================
\section{Results}
\label{sec:results}
% =====================================================================

\subsection{Acceptance and Clarity of the Instrument}

\begin{figure}[htbp]
    \centering
    \includegraphics[width=0.92\linewidth]{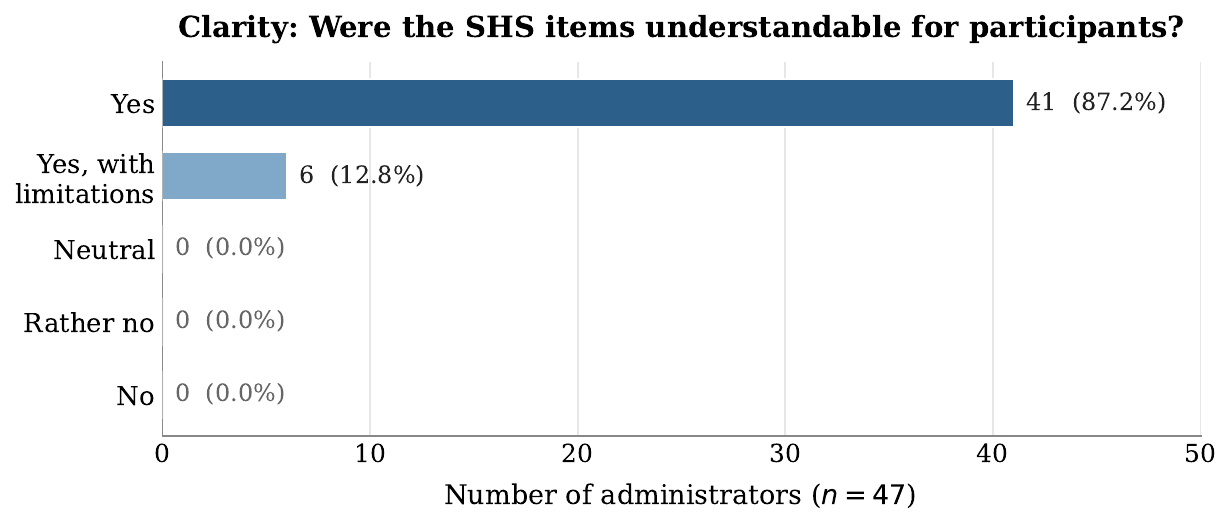}
    \caption{\textbf{Perceived clarity of the SHS items.} Distribution of
    administrators' ratings ($n = 47$) of whether the SHS items were
    understandable for participants. A large majority (87.2\%) reported the
    items were understandable and the remainder (12.8\%) noted only minor
    limitations; no administrator rated the items neutral or unclear.}
    \label{fig:clarity}
\end{figure}

\begin{figure}[htbp]
    \centering
    \includegraphics[width=0.92\linewidth]{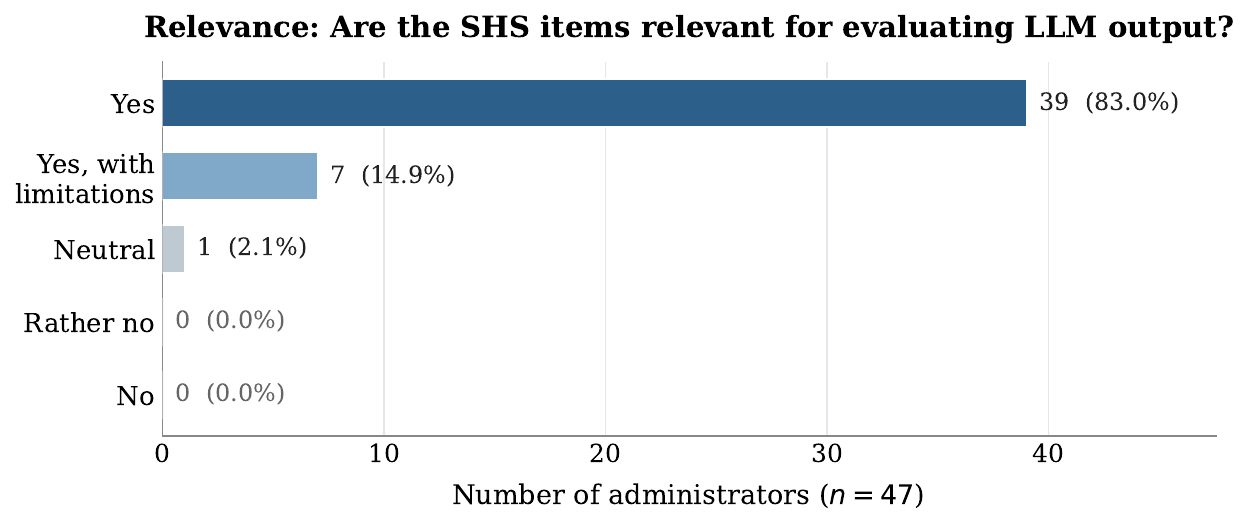}
    \caption{\textbf{Perceived relevance of the SHS items.} Distribution of
    administrators' ratings ($n = 47$) of whether the SHS items are relevant for
    evaluating LLM output. Most (83.0\%) rated them relevant and a further
    14.9\% relevant with limitations; a single rating (2.1\%) was neutral and
    none negative.}
    \label{fig:relevance}
\end{figure}

\begin{figure}[htbp]
    \centering
    \includegraphics[width=0.92\linewidth]{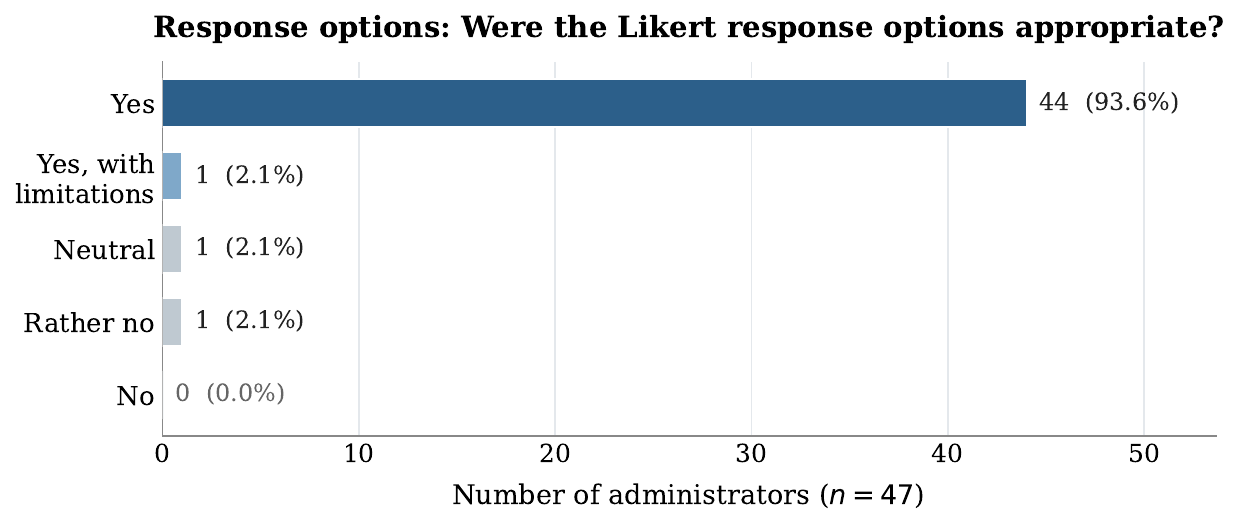}
    \caption{\textbf{Appropriateness of the response options.} Distribution of
    administrators' ratings ($n = 47$) of whether the five-point Likert response
    options were appropriate. The overwhelming majority (93.6\%) found them
    suitable for expressing their judgments, with only isolated reservations.}
    \label{fig:options}
\end{figure}

\begin{figure}[htbp]
    \centering
    \includegraphics[width=0.92\linewidth]{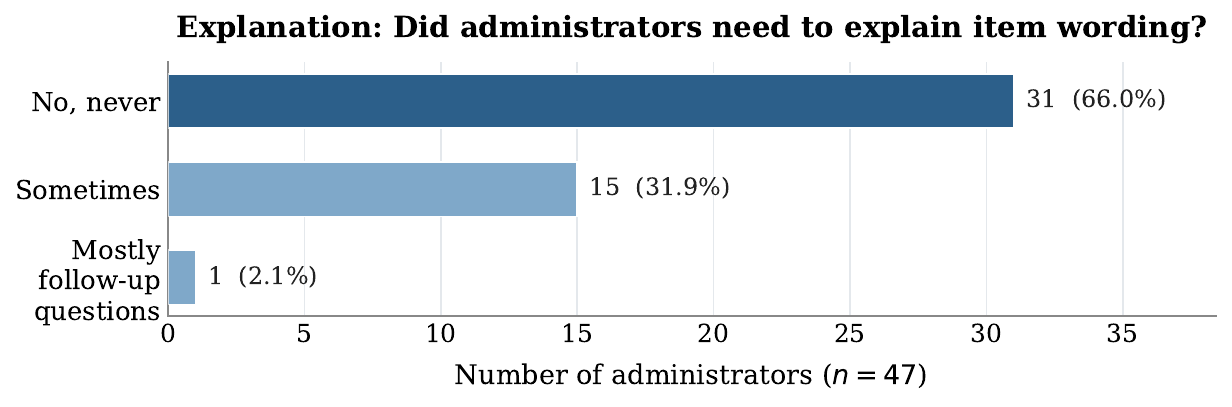}
    \caption{\textbf{Need for additional explanation.} Distribution of
    administrators' reports ($n = 47$) of whether they had to explain item
    wording to participants. Two thirds (66.0\%) never needed to, 31.9\% did so
    occasionally, and only one (2.1\%) needed to do so for most follow-up
    questions, indicating that the items are largely self-explanatory.}
    \label{fig:explanation}
\end{figure}

\begin{figure}[htbp]
    \centering
    \includegraphics[width=0.92\linewidth]{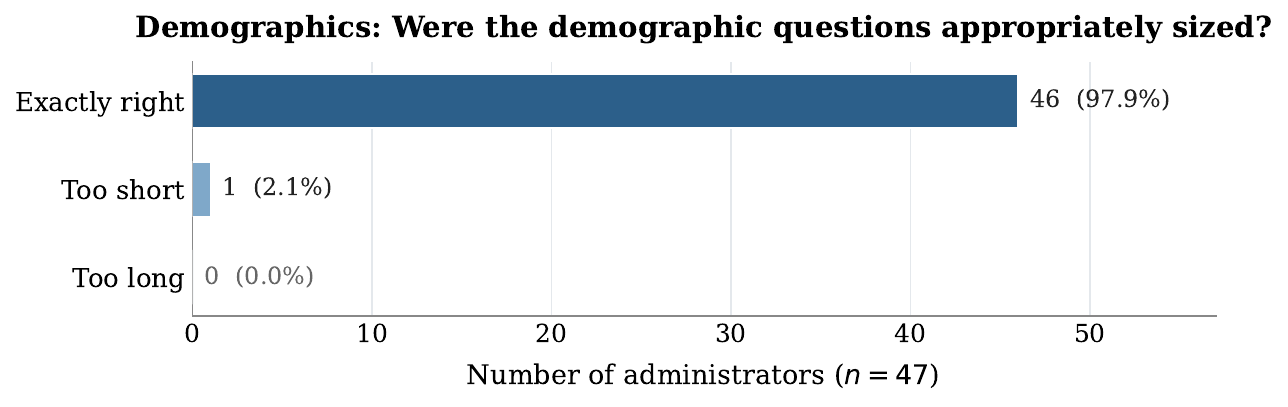}
    \caption{\textbf{Length of the demographic section.} Distribution of
    administrators' ratings ($n = 47$) of the demographic questions. Nearly all
    (97.9\%) judged the section appropriately sized, supporting the overall
    brevity of the protocol.}
    \label{fig:demographics}
\end{figure}

Table~\ref{tab:results_summary} summarizes administrators' judgments of the
instrument. SHS was rated clear (87.2\%), relevant for LLM evaluation (83.0\%),
and appropriately scaled in its response options (93.6\%); most administrators
needed no additional explanation of item wording (66.0\%), and the demographic
section was judged appropriately sized (97.9\%). Detailed response distributions
appear in Figures~\ref{fig:clarity}--\ref{fig:demographics}.

\begin{table}[htbp]
\centering
\caption{Acceptance of the SHS instrument by trained administrators
($n = 47$ experimenters; all percentages are of the 47 administrators).}
\label{tab:results_summary}
\begin{tabular}{lcc}
\toprule
\textbf{Evaluation aspect} & \textbf{Most frequent response} & \textbf{Percentage} \\
\midrule
Clarity of SHS items & Yes & 87.2\% \\
Relevance for LLM evaluation & Yes & 83.0\% \\
Appropriateness of response options & Yes & 93.6\% \\
No explanation required & No, never & 66.0\% \\
Demographic questions length & Exactly right & 97.9\% \\
\bottomrule
\end{tabular}
\end{table}

\subsection{Psychometric Properties}

\paragraph{Internal consistency.}
Cronbach's $\alpha$ across the ten items was $\alpha = .87$ (95\% CI
$[.84, .90]$), exceeding the $.70$ threshold for acceptable reliability and the
$.80$ benchmark for good reliability in applied settings
\citep{Sauro:2016:Quantifying}. Corrected item--total correlations ranged from
$.51$ to $.71$ and no single deletion raised $\alpha$ materially
(Supplementary Material~S4), indicating that every item contributes to a
coherent underlying construct and supporting use of the aggregate score as a
reliable summary.

\paragraph{Internal structure.}
Pearson correlations between the five dimension scores
(Table~\ref{tab:correlations}) were positive and moderate to strong
($r = .42$--$.72$, all $p < .001$). Following effect-size conventions for
correlational evidence in HCI \citep{Cairns:2019:Statistics}, these values
indicate dimensions that are systematically related yet non-redundant: large
enough to share a common perceived-reliability factor, but well below unity,
which would signal that separate dimensions are unwarranted. The strongest
association, between Factual Accuracy and Source Reliability ($r = .72$), is
theoretically expected because both concern the veracity of output; the weakest,
involving Responsiveness to Guidance ($r = .42$--$.51$), is consistent with its
distinct interactional character (controllability rather than correctness).

\begin{table}[htbp]
\centering
\caption{Pearson correlations between SHS dimension scores ($N = 386$
evaluations);
all coefficients significant at $p < .001$.}
\label{tab:correlations}
\begin{tabular}{lccccc}
\toprule
& \textbf{FA} & \textbf{SR} & \textbf{LC} & \textbf{DP} & \textbf{RG} \\
\midrule
Factual Accuracy (FA)    & 1.00 &      &      &      &      \\
Source Reliability (SR)  & 0.72 & 1.00 &      &      &      \\
Logical Coherence (LC)   & 0.68 & 0.61 & 1.00 &      &      \\
Deceptiveness (DP)       & 0.54 & 0.49 & 0.57 & 1.00 &      \\
Responsiveness (RG)      & 0.45 & 0.42 & 0.51 & 0.48 & 1.00 \\
\bottomrule
\end{tabular}
\end{table}

\paragraph{Paired-item validity.}
After reverse-scoring the negatively worded items, the within-dimension
correlation between each positive/negative pair was strong (Factual Accuracy
$r = .79$; Logical Coherence $r = .74$; Source Reliability $r = .71$;
Responsiveness $r = .68$; Deceptiveness $r = .65$; all $p < .001$). Convergence
of the two oppositely worded items of each dimension is the expected signature
of a valid bipolar design and indicates that participants engaged with item
meaning rather than responding to surface polarity---the acquiescence safeguard
that motivates alternating wording in SUS and SCS
\citep{Brooke:1996:SUS,Lewis:2018:SUS}.

\paragraph{Response behavior.}
To check that ratings reflected graded judgment rather than satisficing, we
tested the marginal distribution of responses against a uniform null with a
chi-square goodness-of-fit test. Responses departed strongly from uniformity
($\chi^2(4) = 187.3$, $p < .001$), and item-level distributions
(Supplementary Material~S3) show participants using the full width of the scale
rather than defaulting to its endpoints---behavior consistent with engaged,
discriminating evaluation rather than response sets
\citep{Cairns:2019:Statistics}. Overall SHS scores were approximately normally
distributed (Shapiro--Wilk $W = .987$, $p = .051$; skewness $-0.18$, kurtosis
$-0.34$), satisfying the distributional assumptions of the parametric analyses
reported above.

\subsection{The Observed Dimension Profile Is Diagnostic}
\label{sec:profile}

\begin{figure}[htbp]
    \centering
    \includegraphics[width=0.95\linewidth]{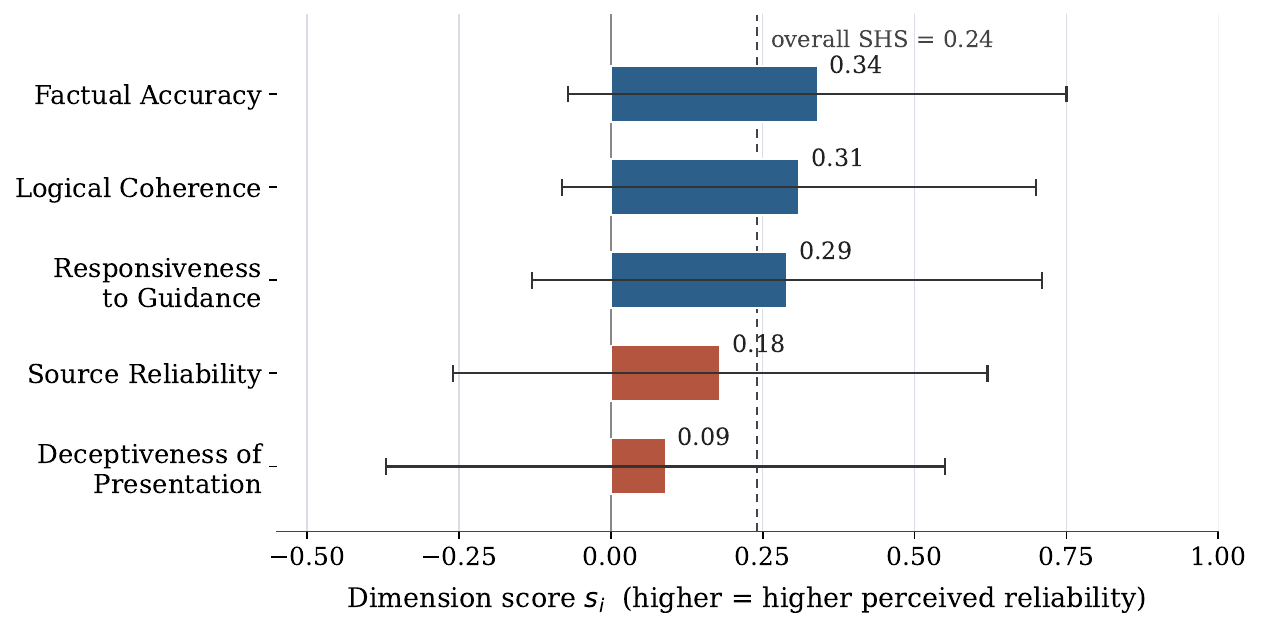}
    \caption{\textbf{SHS dimension profile under the hallucination-eliciting
    prompt protocol} ($N = 386$ evaluations). Bars show mean dimension scores $s_i$ on the
    symmetric $[-1,+1]$ scale; whiskers are $\pm 1$ SD; the dashed line marks
    the overall SHS mean of $0.24$. Source Reliability and Deceptiveness of
    Presentation (darker bars) fall well below the other three dimensions,
    locating the perceived weakness in traceability and framing rather than in
    raw correctness. Higher is better throughout.}
    \label{fig:profile}
\end{figure}

Beyond establishing that the instrument behaves well, the study yields one
substantive empirical result about how the systems under test were perceived,
and it illustrates why a five-dimensional instrument is worth its extra items.
Figure~\ref{fig:profile} shows the mean dimension scores
(Supplementary Material~S3, Table~S3).

Under the common two-category prompt protocol the five dimensions separated
clearly.
Factual Accuracy ($M = 0.34$, $SD = 0.41$), Logical Coherence ($M = 0.31$,
$SD = 0.39$), and Responsiveness to Guidance ($M = 0.29$, $SD = 0.42$) sat
near or above the overall mean of $0.24$, whereas Source Reliability
($M = 0.18$, $SD = 0.44$) and above all Deceptiveness of Presentation
($M = 0.09$, $SD = 0.46$) were markedly lower. The item-level distributions
locate the effect precisely: in 34.3\% of evaluations the rater agreed or
strongly agreed that sources were omitted or invented (Q4), and in 43.8\% that
false information was presented confidently and misleadingly (Q8), against
22.8\% agreeing that false or fabricated information was generated frequently
(Q2).

Read as a profile rather than as a single number, this says something an
aggregate score cannot. Participants judged that the systems were, under
adversarial prompting, moderately accurate and moderately coherent, but that
what they could not do was \emph{check} the output or \emph{tell from its
presentation} when it was wrong. The perceived weakness is in traceability and
framing, not primarily in raw correctness. Deceptiveness also carried the
largest dispersion ($SD = 0.46$) and the lowest proportion of internally
consistent paired judgments (61.0\% with $|c_i| \leq 0.25$, versus 74.3\% for
Logical Coherence; Supplementary Material~S3), which is itself informative:
whether a confident falsehood is recognizable is the judgment users find
hardest to make, and the one on which they least agree with themselves.

We are explicit about what this result is and is not. It is a finding about
\emph{perception} under a stress-test protocol, pooled across the systems in
Table~\ref{tab:systems}; it is not a claim that these systems fabricate sources
at any particular rate, and the absolute levels would very likely be higher
under ordinary use. Its value is that it demonstrates the instrument doing the
work it was designed for---separating failure modes that a single hallucination
score would have merged---and that it identifies source traceability and
non-deceptive presentation as the leverage points a designer would act on
first. That the two weakest dimensions are precisely those that automated
factuality benchmarks do not measure would, \emph{if} criterion validity were
established, be the strongest argument for the instrument's complementary
value. It is not itself evidence that the perception is correct, and we draw no
conclusion from it here; the question is posed formally in
Section~\ref{sec:criterion} and the joint-use proposition in
Section~\ref{sec:joint}.

\subsection{Feasibility and Diagnostic Value}

SHS was fast to administer (mean completion 4.2 min, $SD = 1.8$) and did not
disrupt interaction. The paired-item structure provided a useful diagnostic:
similar ratings on both items of a dimension typically reflected genuine
uncertainty or mixed impressions rather than random responding, offering a
simple quality-control signal. Beyond feasibility, administrators reported that
SHS helped participants \emph{articulate why} an output felt
unreliable---distinguishing, for example, an obviously wrong statement from a
coherent but ungrounded one---capturing aspects of perceived reliability not
reducible to string-based metrics. We note that this last observation is
administrator testimony collected in the feedback questionnaire, not a measured
outcome.

% =====================================================================
\section{Validity Evidence: Established and Outstanding}
\label{sec:validity}
% =====================================================================

Scale validation is cumulative, and no single study completes it
\citep{Cronbach:1955:ConstructValidity,Messick:1995:Validity}. We therefore
separate what the present study establishes from what it does not.

\subsection{Evidence Established Here}

\begin{itemize}[leftmargin=*,itemsep=2pt]
\item \textbf{Content evidence.} Dimensions were derived from the documented
failure-mode taxonomy of Section~\ref{sec:background} and from user-reported
hallucination categories in deployment \citep{Massenon:2025:UserReported}; item
clarity and relevance were independently judged acceptable by 87.2\% and 83.0\%
of trained administrators.
\item \textbf{Internal-structure evidence.} $\alpha = .87$; item--total
correlations $.51$--$.71$; inter-dimension correlations $.42$--$.72$, related
but non-redundant; positive/negative item pairs converging at $.65$--$.79$
after reverse scoring.
\item \textbf{Response-process evidence.} Departure from uniform responding,
full use of the scale range, and paired-item coherence in 61--74\% of
judgments, indicating engagement with item meaning rather than response sets.
\end{itemize}

\subsection{Criterion Validity: Not Established, and How to Establish It}
\label{sec:criterion}

The central limitation of this work is that \textbf{SHS has not been correlated
with any objective measure of hallucination}. Every coefficient reported in
Section~\ref{sec:results} concerns the internal behavior of the instrument.
None of them shows that perceived reliability tracks actual reliability, and
the literature reviewed in Section~\ref{sec:background} gives concrete reasons
to expect imperfect tracking in both directions: users over-rely on confident
output \citep{Si:2024:Verify}, are influenced by surface cues that are
uninformative about faithfulness \citep{Chiesurin:2023:Dangers}, and shift their
judgments in response to expressed uncertainty independently of accuracy
\citep{Kim:2024:Uncertainty}. A plausible prior is therefore a positive but
moderate association, with systematic divergence concentrated in the
Deceptiveness dimension---which is precisely where divergence would matter
most.

We regard this as the necessary next study rather than as a caveat, and we
specify it so that it can be executed by us or by others:

\begin{enumerate}[label=(C\arabic*),leftmargin=*,itemsep=3pt]
\item \textbf{Paired design.} Collect $N \geq 200$ human--LLM interactions
under the protocol of Section~\ref{sec:protocol} with full transcript logging,
each rated with SHS by the participant.
\item \textbf{Objective annotation of the same transcripts.} Score each
transcript on three independent objective criteria: (i) expert atomic-fact
annotation in the manner of FActScore \citep{Min:2023:FActScore}, giving a
factual-precision rate; (ii) an automated reference-free detector, semantic
entropy \citep{Farquhar:2024:SemanticEntropy} and SelfCheckGPT
\citep{Manakul:2023:SelfCheckGPT}, giving a model-side uncertainty signal;
(iii) span-level citation-support annotation in the manner of RAGTruth
\citep{Niu:2024:RAGTruth} for the source-request prompts.
\item \textbf{Pre-registered hypotheses.} $H_1$: overall SHS correlates
positively with factual-precision rate, $r \geq .40$. $H_2$: the Source
Reliability dimension correlates more strongly with citation-support rate than
does any other dimension (discriminant prediction). $H_3$: the residual of SHS
on factual precision is predicted by the Deceptiveness dimension---that is,
users systematically over-rate systems whose errors are confidently
framed. $H_3$ is the diagnostically important hypothesis, because confirming it
would show that SHS measures something automated metrics cannot, and
disconfirming it would show that SHS is partly redundant with them.
\item \textbf{Falsification criterion.} If overall SHS fails to correlate with
factual precision at $r \geq .20$, the instrument does not measure what its
name implies and should be renamed or restructured.
\end{enumerate}

Stating a falsification criterion is deliberate. An instrument that cannot in
principle be shown wrong is not a measurement instrument.

\subsection{Other Outstanding Evidence}
\label{sec:outstanding}

Convergent and discriminant validity against neighbouring subjective
constructs---trust in automation \citep{Lee:2004:TrustAutomation}, perceived
usability \citep{Brooke:1996:SUS}, perceived explanation quality
\citep{Holzinger:2020:Causability}---has not been assessed; SHS should
correlate moderately with all three and be distinguishable from each.
Test--retest stability, measurement invariance across
languages and demographic subgroups, and confirmatory factor analysis of the
five-dimension structure on an independent sample all remain to be done. The
present study is a first validation, not a completed validation programme.

% =====================================================================
\section{Positioning SHS Among Existing Evaluation Systems}
\label{sec:positioning}
% =====================================================================

This section situates SHS in the evaluation landscape. It is a conceptual
analysis, not an empirical result, and we present it as such.

\subsection{The Landscape}

Table~\ref{tab:systems_comparison} contrasts representative systems on
technical versus user-experience focus, automation, scale, and dimensionality.
The landscape is dominated by automation-centric, largely single-dimensional,
offline benchmarks that encode expert definitions of error. Four gaps recur: a
binary-versus-multidimensional gap, an offline-versus-interactive gap, an
expert-versus-end-user gap, and a missing \emph{guidance} dimension---whether
users can steer the system back toward accuracy, which no automated benchmark
measures because it is a property of an exchange rather than of a response.
Holistic evaluation efforts \citep{Bommasani:2023:HolisticEvaluation} and
interaction-centric studies of agentic systems
\citep{Park:2023:GenerativeAgents} likewise argue that task-level accuracy
alone is insufficient. SHS occupies the user-experience, multi-dimensional,
interaction-aware cell of this space, requiring no ground truth and no model
access.

\begin{table}[htbp]
\centering
\caption{Hallucination evaluation systems: technical versus user-experience
focus. $\bullet$ = primary focus; $\circ$ = secondary/not addressed.}
\label{tab:systems_comparison}
\small
\begin{tabular}{>{\raggedright\arraybackslash}p{2.7cm}ccccc>{\raggedright\arraybackslash}p{2.9cm}}
\toprule
\textbf{System} & \textbf{Tech.} & \textbf{UX} & \textbf{Auto.} & \textbf{Human} & \textbf{Dims.} & \textbf{Primary focus} \\
\midrule
TruthfulQA        & $\bullet$ & $\circ$ & $\bullet$ & $\circ$ & 1 & Truthfulness \\
HaluEval          & $\bullet$ & $\circ$ & $\bullet$ & $\circ$ & 1 & Binary detection \\
FEVER             & $\bullet$ & $\circ$ & $\bullet$ & $\circ$ & 1 & Claim verification \\
FActScore         & $\bullet$ & $\circ$ & $\bullet$ & $\circ$ & 1 & Atomic precision \\
SelfCheckGPT      & $\bullet$ & $\circ$ & $\bullet$ & $\circ$ & 1 & Consistency-based \\
Semantic entropy  & $\bullet$ & $\circ$ & $\bullet$ & $\circ$ & 1 & Meaning-level uncertainty \\
RAGTruth          & $\bullet$ & $\circ$ & $\bullet$ & $\bullet$ & 1 & Span-level RAG errors \\
HHEM              & $\bullet$ & $\circ$ & $\bullet$ & $\circ$ & 1 & Summ. faithfulness \\
RAGAS             & $\bullet$ & $\circ$ & $\bullet$ & $\circ$ & 2 & RAG faithfulness \\
LLM-as-judge      & $\bullet$ & $\circ$ & $\bullet$ & $\circ$ & Flex. & Automated scoring \\
Human annotation  & $\bullet$ & $\bullet$ & $\circ$ & $\bullet$ & Flex. & Gold-standard \\
User reviews      & $\circ$ & $\bullet$ & $\circ$ & $\bullet$ & --- & In-the-wild perception \\
\midrule
\textbf{SHS (ours)} & $\circ$ & $\bullet$ & $\circ$ & $\bullet$ & 5 & Perceived reliability \\
\bottomrule
\end{tabular}
\end{table}

\subsection{A Protocol for Joint Use---and How to Test Whether It Helps}
\label{sec:joint}

\paragraph{Protocol.} For an interactive LLM system under evaluation:
(1)~run the applicable automated factuality measures on a fixed task set,
yielding a system-side score; (2)~run a fixed interaction protocol with
$\geq 30$ representative users, collecting SHS per session; (3)~report the
automated score, the overall SHS, and the five-dimension profile together;
(4)~treat divergence between them as the object of interest---in particular, a
high automated score paired with a low Deceptiveness dimension score would, on
the proposition below, point to a system that is accurate but whose errors are
hard for users to catch, and would call for interface work (uncertainty
display, citation surfacing) rather than model work.

\paragraph{Proposition, not result.} The claim that this pairing yields
information neither component provides alone is a hypothesis. It is tested by
$H_3$ of Section~\ref{sec:criterion}: if the SHS residual on objective accuracy
is systematically predicted by presentation properties, the pairing is
informative; if not, SHS adds little beyond the automated score and the
protocol should be abandoned. We label this a proposition throughout and draw
no conclusions from it in this paper.

% =====================================================================
\section{SHS in the Family of Human-Centered Scales}
\label{sec:family}
% =====================================================================

SHS extends a coherent family of compact human-centered instruments. SUS
captures perceived ease of use and learnability
\citep{Brooke:1996:SUS,Lewis:2018:SUS,Bangor:2008:Empirical}; SCS captures
perceived transparency and understandability of AI reasoning
\citep{Holzinger:2020:Causability}; SHS captures perceived reliability. All
three share balanced item polarity, a compact ten-item format completable in
minutes, and domain-agnostic applicability
(Table~\ref{tab:scale_comparison}). SHS adds three structural innovations: an
explicit five-dimensional structure that localizes specific failure modes
rather than a single aggregate; a built-in consistency diagnostic from the
paired-item design; and a symmetric $[-1,+1]$ score with intuitive sign
semantics that remains rescalable to conventional ranges. For comprehensive
evaluation of interactive LLM systems---especially in high-stakes,
accessibility-sensitive deployment contexts
\citep{Stephanidis:1999:InformationSociety,Shneiderman:2020:HCAI}---we suggest
administering SUS, SCS, and SHS together to obtain a view across usability,
explainability, and reliability; the discriminant validity of the three
instruments against one another remains to be established
(Section~\ref{sec:validity}).

\begin{table}[htbp]
\centering
\caption{Structural comparison of SHS with established human-centered scales.}
\label{tab:scale_comparison}
\small
\renewcommand{\arraystretch}{1.15}
\begin{tabular}{@{}l>{\raggedright\arraybackslash}p{3.1cm}>{\raggedright\arraybackslash}p{3.1cm}>{\raggedright\arraybackslash}p{3.6cm}@{}}
\toprule
\textbf{Characteristic} & \textbf{SUS} & \textbf{SCS} & \textbf{SHS} \\
\midrule
Number of items        & 10 & 10 & 10 \\
Response scale         & 5-pt Likert & 5-pt Likert & 5-pt Likert \\
Item structure         & Alternating $\pm$ & Alternating $\pm$ & Paired $\pm$ per dimension \\
Score range (native)   & 0--100 & 0--100 & $[-1, +1]$ \\
Score range (rescaled) & --- & --- & 0--100 \\
Dimensions             & 1 & 1 & 5 \\
Consistency diagnostic & No & No & Yes \\
Target construct       & Perceived usability & Perceived causability & Perceived reliability \\
\bottomrule
\end{tabular}
\end{table}

% =====================================================================
\section{Strengths, Limitations, and Ethical Considerations}
\label{sec:limitations}
% =====================================================================

\subsection{Strengths}

SHS can be applied with minimal instruction, completes in about four minutes,
and requires no ground truth, reference corpus, or model access. Its
five-dimensional structure distinguishes manifestations of perceived
unreliability that a single score merges, as Section~\ref{sec:profile}
demonstrates, and the paired-item design supplies an internal diagnostic signal
useful for quality control. Instrument, scoring code, and a multilingual
calculator are openly released.

\subsection{Limitations}

\paragraph{L1: No criterion validity} (non-claims N2, N5)\textbf{.} SHS has not
been correlated with any objective hallucination measure. This is the principal
limitation; see Section~\ref{sec:criterion} for the study that would address
it. It is also why non-claim N1 of Section~\ref{sec:nonclaims} holds: an
instrument with no criterion evidence cannot be a detector.

\paragraph{L2: The protocol is neither naturalistic nor experimentally
controlled} (non-claim N4)\textbf{.} The study used a supervised,
hallucination-eliciting protocol, so absolute score levels do not estimate
perceived reliability in ordinary use, and the reliability coefficients were
obtained under variance conditions more favorable than naturalistic use would
provide (Section~\ref{sec:ecological}). At the same time, because participants
composed their own prompts, the stimulus was not held constant and the design
does not meet the standard of a controlled experiment. It occupies an
intermediate position, and we claim the evidentiary reach of neither endpoint.

\paragraph{L3: The stimulus is participant-generated.} Prompt content varied
across participants by design (Section~\ref{sec:protocol}). This was chosen so
that participants could exploit their own domain knowledge, where subtle
hallucinations are detectable, but it means the observed score distribution
reflects a heterogeneous stimulus set rather than a fixed one, and that exact
replication requires the deposited prompt corpus rather than a reprinted
battery.

\paragraph{L4: Pooling across systems} (non-claim N3)\textbf{.} Validation
statistics pool across the
systems in Table~\ref{tab:systems}. Between-system differences in hallucination
behavior enter as stimulus variance, so the reported inter-dimension
correlations characterize the instrument across a heterogeneous stimulus set
rather than for a fixed system. Re-estimation within a single system is needed
to separate participant disagreement from stimulus variation.

\paragraph{L5: Sample composition} (non-claim N6)\textbf{.} The sample is a
convenience sample of
German- or English-speaking adults without AI/ML professional experience. We
make no claim of demographic diversity, and measurement invariance across
subgroups and languages is untested.

\paragraph{L6: Subjectivity is intrinsic, not incidental.} SHS depends on the
rater's background knowledge, attention, and interpretation of context, and can
be influenced by framing or prompting strategy. A rater who cannot detect an
error cannot report it. The paired-item structure mitigates but does not
eliminate this; SHS provides no ground-truth correctness or certification.

\paragraph{L7: Self-report about the instrument.} Clarity and relevance figures
are administrator self-reports, subject to the usual acquiescence and
demand-characteristic pressures of feedback questionnaires.

\paragraph{L8: Single-occasion measurement.} No test--retest data are
available, so temporal stability is unknown.

Accordingly, scores are best read as \emph{relative} indicators supporting
comparison and monitoring across tasks, contexts, and iterative system changes,
rather than as absolute measures of model reliability.

\subsection{Ethical Considerations}

Two risks deserve naming. First, a favorable SHS score could be used as
marketing evidence that a system is reliable; the interpretation bands
(Table~\ref{tab:interpretation}) and the non-claims of
Section~\ref{sec:nonclaims} are written to make such use plainly unsupported,
and we recommend that SHS scores never be reported without the accompanying
prompting regime and rater population. Second, an instrument that measures
perceived reliability could in principle be optimized against---a system tuned
to appear reliable rather than to be reliable, which the sycophancy literature
shows is an attainable and rewarded objective
\citep{Sharma:2024:Sycophancy}. This is a further reason SHS must be paired
with objective measurement rather than substituted for it. Consistent with
human-centered AI principles \citep{Shneiderman:2020:HCAI}, SHS is intended to
keep the human in the loop---as the arbiter of reliability---rather than to
automate that judgment away.

% =====================================================================
\section{Conclusion and Future Work}
% =====================================================================

We introduced the System Hallucination Scale (SHS), a lightweight instrument
that makes the \emph{perceived} reliability of LLM output measurable along five
interpretable dimensions: factual accuracy, source reliability, logical
coherence, deceptiveness of presentation, and responsiveness to user guidance.
A supervised validation study with 193 participants (386 LLM evaluations) and 47 trained
administrators showed that SHS is clear (87.2\%), fast (mean 4.2 min), and
internally sound ($\alpha = .87$), with an internal structure whose dimensions are related but
non-redundant. Under a hallucination-eliciting protocol the observed dimension
profile separated cleanly, locating the perceived weakness of the systems under
test in source traceability and non-deceptive presentation rather than in raw
correctness---a distinction that a single hallucination score cannot express.

We are equally clear about what has not been shown. SHS has not been correlated
with any objective hallucination measure, the protocol was supervised and
adversarially oriented rather than naturalistic, and the sample does not license claims of population generality.
Section~\ref{sec:criterion} specifies a pre-registered criterion-validity study
with an explicit falsification criterion; that study, not further instrument
refinement, is the appropriate next step. Beyond it, we intend to establish
convergent and discriminant validity against SUS, SCS, and trust-in-automation
measures, test measurement invariance across the three released languages and
across demographic subgroups, examine robustness under naturalistic
self-directed interaction, and study longitudinal change as systems are
updated.

To support reproducibility and adoption, all items, scoring logic, reference
code, the multilingual calculator, and evaluation materials are openly
available (Supplementary Material~S1).

% =====================================================================
\section*{Declarations}
\addcontentsline{toc}{section}{Declarations}
% =====================================================================

\noindent\textbf{Ethical approval.} All experimental protocols were approved by
the Ethics Committee of the Medical University of Graz (EK-Number 34-527 ex
21/22), and all methods were carried out in accordance with relevant guidelines
and regulations.

\noindent\textbf{Informed consent.} Informed consent was obtained from all
participants. Participation was voluntary, fully anonymous, and for scientific
purposes only.

\noindent\textbf{Disclosure statement.} The authors report no competing
interests.

\noindent\textbf{Author contributions.} A.H.\ and H.M.\ developed the scale;
A.H.\ conducted the empirical study; H.M.\ analyzed the results; D.S.\ and M.P.\
provided feedback; all authors contributed to writing.

\noindent\textbf{Data availability.} All evaluation materials, questionnaires,
and anonymized response data are available in the Supplementary Material and the
project repository: \url{https://github.com/human-centered-ai-lab/system-hallucination-scale}.

\noindent\textbf{Funding.} This research was funded in part by the Austrian
Science Fund (FWF), Project ``Explainable AI'', Grant P-32554, and in part by
the European Union's Horizon~2020 research and innovation programme ``AI-powered
Data Curation and Publishing Virtual Assistant (AIDAVA)'', module ``System
Hallucination Scale (SHS)'', under grant agreement No.~101057062. This
publication reflects only the authors' view; the European Commission is not
responsible for any use that may be made of the information it contains.

\noindent\textbf{Generative AI statement.} In accordance with the Taylor \&
Francis policy on the use of AI-assisted technologies, the authors disclose the
following. AI-assisted language-editing tools (Grammarly; DeepL) were used for
grammar, spelling and language revision. AI-assisted search tools (Perplexity;
Google AI) were used to support literature discovery, and every source so
identified was subsequently verified against the publisher record. A large
language model assistant (Anthropic Claude) was used during the preparation of
this revision to support drafting and copy-editing of manuscript text and to
cross-check bibliographic metadata. No generative AI system is listed as an
author; no generative AI was used to generate research data, results, figures
or other artwork; and all AI-assisted output was reviewed and edited by the
authors, who take full responsibility for the content of this article.

% =====================================================================
\bibliography{clean_v1}
% =====================================================================

\end{document}